# Image Classification base on PCA of Multi-view Deep Representation

Yaoqi Sun[1], Liang Li[2*], Liang Zheng[1*], Ji Hu[1], Yatong Jiang[1], Chenggang Yan[1]

[1] Hangzhou Dianzi University, Hanzhou, China

[2] Institute of Computing Technology, Chinese Academy of Sciences, Kexueyuan South Road #6, Haidian District, Beijing, China

* Corresponding author. E-mail address: zhengliang@hdu.edu.cn（Liang Zheng）

* Corresponding author. E-mail address: liang.li@ict.ac.cn (Liang Li).

**Abstract:** In the age of information explosion, image classification is the key technology of dealing with and organizing a large number of image data. Currently, the classical image classification algorithms are mostly based on RGB images or grayscale images, and fail to make good use of the depth information about objects or scenes. The depth information in the images has a strong complementary effect, which can enhance the classification accuracy significantly. In this paper, we propose an image classification technology using principal component analysis based on multi-view depth characters. In detail, firstly, the depth image of the original image is estimated; secondly, depth characters are extracted from the RGB views and the depth view separately, and then the reducing dimension operation through the PCA is implemented. Eventually, the SVM is applied to image classification. The experimental results show that the method has good performance.

**Keyword**：Image classification；Principal component analysis；multi-view depth characters

## 1. Introduction

Today, as the society has entered the information era, in addition to the large amount of text information, the multimedia information (audio, image, video, etc.) also presents explosive increase [1].

It is a serious problem of choosing the image information people need from the vast amount of image data. On the one hand, people want to obtain more data for more comprehensive information; on the other hand, it is more difficult to accurately and quickly obtain image information from more images. Thus, how to effectively organize and manage the disorder image data, and to find images we need from it accurately, comprehensively and quickly, needs to be done urgently. Image classification is the key technology of handling and organizing a large number of image data, which can solve the problem of image data disorder.

Feature extraction is the key step in the image classification, separated from the computer vision and image operation, when using computer to analyse and deal with the image information, determine the invariant feature of images, and then extract features for solving practical problems. Feature extraction techniques have been applied to all areas of our life, such as ancient architecture reconstruction and protection [2], remote sensing image analysis [3], urban planning and medical diagnosis [4]. However, it is still one of the difficulties and hot spots in the field of image operation when concerning extracting the image features of strong expression ability and anti-noise ability. Color, texture and shape features are the basic lower-level features of the image. The color feature has globality, which can be extracted by the color histogram, color set, color moment and so on [5]. It can simply describe the proportions of different colors in the whole image. The color feature is ideal for describing images that are difficult to automatically split, and the distribution of space shouldn't be considered [6]. However, it cannot describe the local distribution in the image and the description concerning the space position of various colors in the image [7]. The texture feature is similar to the color feature, and it is also a characteristic of globality. Ceryan and Jain [8] sum texture feature extraction methods up into five categories: structural methods, signal operation methods, geometric methods, model methods and statistical methods. When comparing color feature extraction, it can be found that texture features will not match for some local deviations, when texture features at the same time have the excellent rotation invariance and good resistance to noise interference [9-10]. However, when the pixel resolution change of the image is obvious, the deviation of the texture feature will increase obviously [11]. There are two methods of shape feature extraction. To be specific, one is the regional feature which mainly focuses on the whole shape region of the image; the other is the contour feature, which is aimed at the outer boundary of the object. The predecessors have presented many typical shape feature extraction methods: boundary eigenvalue methods (the outer boundary of the image) [12], geometric parameter methods (image geometric parameterization operation) [13], shape invariant moment methods (finding image moment invariant features) [14], Fourier shape description methods (Fourier transform methods) [15], etc. Its advantage is the overall grasp of the image target. If the target in the image is deformed, the stability of the description will be reduced to a large extent [16]. Meanwhile, due to the globality of the shape feature, the space requirements of the calculation time and storage are relatively high.

Any method of image feature extraction has its advantages and insuperable defects due to its inherent characteristics. In addition, most of the existing algorithms focus on the analysis from a single feature, while ignoring the correlation factors between the characteristics of the image. If the multi-method with the multi-feature fusion strategy can be adopted, the capacity of image classification should be enhanced significantly.

The current classical image classification algorithms are mainly based on RGB images or grayscale images, and do not make good use of the depth information of objects or scenes. The Spatial Pyramid Matching (SPM) framework has overcome the lost space information in the BOF algorithm and effectively improved the accuracy of image classification [17]. However, the obtained images must be based on nonlinear kernel functions, such as the nonlinear SVM (Support Vector Machine, SVM), to obtain better classification models, and thus, the efficiency of image classification is unsatisfying. The images obtained by the computers are usually affected by the background, brightness and view point change. In this case, the image classification becomes a challenging problem in the fields of computer vision and artificial intelligence.

With the advent of the unsupervised image depth estimation, the research on the rgb-d image has rapidly become one of the research hotspots in the field of computer vision [18, 19]. The RGB image is a color image containing the color information in the scene. The depth image is a grayscale image, and the pixel value denotes the relative distance in the scene. Color images have good detail texture and color information, but are susceptible to environmental factors such as illumination. However, the depth image can obtain more reliable geometric information without the disturbance of the light shadow and the texture on the surface of the object. The information contained in these two images has a strong complementary effect. The depth image is featured with robustness for the interference of objects whose colors are similar to the background and lighting changes, and its edge can accurately represent the boundary of the object [21]. However, when the object is close to the background, the discrimination ability of the depth information drops sharply, while the color information has a good robustness to the change of the distance. Besides, the RGB-D image using two-dimensional image information represents 3D scene information, bridges the gap between two-dimensional plane and three-dimensional space, and provides the possibility of dealing with three-dimensional problems.

In this paper, we propose an image classification technology using principal component analysis based on multi-view depth characters. In detail, we first estimate the depth image of the original image, and extract depth character from RGB views and the depth view separately. After reducing dimension operation by PCA method, the SVM is adopted for image classification

## 2. Related Work

The content of image features includes color, texture, shape and other visual features. Image feature extraction is the premise of image analysis and image recognition. It is the most effective way of simplifying the expression of high dimensional image data. Local features can only reflect the local characteristics of the image, so it is suitable for image matching and retrieval, which is not suitable for image understanding. The latter is more concerned with global features, such as color texture and shape features, which are vulnerable to environmental interference, such as lighting, rotation, noise and other adverse factors.

In the past 20 years, researches on image feature representation have achieved a lot at home and abroad. In 2004, Lowe [1] proposed the effective Scale Invariant Feature Transform (SIFT) algorithm, when using the original image with the Gaussian convolution of nuclear to establish scale space, which extracts scale invariance feature points based on Gaussian pyramid. In 2006, Bay and Ess [2], based on the idea of the SIFT algorithm, proposed the Speeded Up Robust Features (SURF). The suggested method adopts the approximate Harr wavelet method to extract feature points, while overcoming the disadvantages of large computation cost and slow speed. The HOG algorithm [3] proposed in 2005 extracts features by calculating the gradient directional histogram concerning the local region of the image which shows its excellent properties in in pedestrian detection. Besides, it has been broadly applied to the areas of image recognition and image analysis in combination with the SVM technology. So far, a variety of feature descriptions have been introduced, and the more representative ones are floating-point feature descriptions and binary string feature descriptions. As in the SIFT and SURF algorithms, the feature descriptions using the gradient statistical histogram are the floating-point feature descriptions which are computationally complex and inefficient. Therefore, there are many new feature description algorithms, such as BRIEF. Many of the subsequent binary string feature descriptions such as ORB [4], BRISK [5], and FREAK [6], are improvements on the top of it.

Traditional image feature extraction views are labor-intensive feature projects depending on hand-crafted local descriptors. In recent years, the neural convolution network technology has played an increasingly greater role in the field of image classification. In ILSVRC 2012, the deep convolutional neural network (CNN) has received a lot of attention from researchers for its good performance [12]. It is able to automatically extract image features from a large number of image data and classify them. Compared with traditional feature extraction, the convolution neural network has better performance without the need for artificial image features [7]. However, this novel method also faces greater test complexity and training complexity, as well as a large amount of training time [8] [13]. To deal with this problem, [9] is proposed to move to a new target dataset which two new adaptation layers are learned in, using a pre-trained DCNN model. Besides, this method reduces training time and training data, but improves test complexity. There are other ways of trying to integrate traditional methods and DCNN methods. In [10]'s method, the deep factor of DCNNs is incorporated into the traditional SIFT/fv scheme. Sydorov et al. [11] pointed out that the standard FV aggregator is used as the deep architecture which is replaced by supervised versions. The above methods adopt the deep characteristic of DCNN.

Principal component analysis (PCA) is a useful tool for data compression and information extraction based on overall information [14] [15]. It can convert a number of raw indicators into a few comprehensive indexes without loss of information. Each of these principal components is a linear combination of the original indices, which are not related to each other. That is why the winner has more superior performance than the original index. As a commonly-used statistical way, PCA has been applied widely to pattern recognition and image operation. Besides, many traditional classification methods, such as SVM, have extremely efficient performance in classification tasks with a small training sample size. There are signs in [16]'method that the SVM classifier gives the best sensitivity and specificity results compared with Decision Tree, Random Forest and Naive Bayes classifiers. In general, the Deep CNN model is suitable for feature extractors, and its classification performance is lower than the classical model when the SVM, a traditional classification method can fill in gaps [17].

In our method, the deep feature is extracted from the RGB views and depth views separately, when the high dipartite degree appears, leading to the higher accuracy in image classification. Second, the reducing dimension operation is implemented through the classical PCA. Finally, the popular SVM is employed for image classification. Experiment comparison reveals the promising performance of our proposed approach.

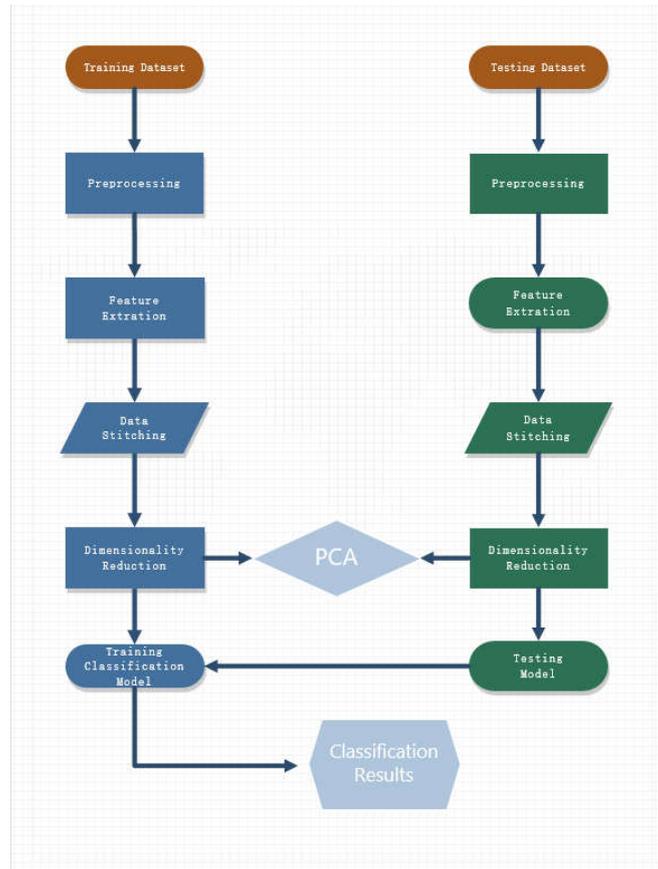

Fig. 1: Our main architecture of the whole paper

## 3. Our Method

Firstly, the training dataset is processed by computers. Or in other words, the sample color picture is divided into three pictures of R, G and B views, and then the convolutional neural network is adopted to extract the features of R, G and B respectively. At the same time, a computer is used to convert the sample color image into a grayscale image, and then the convolutional neural network is employed to extract the feature grayscale image of the sample image. Then, the image is obtained from the R, G and B component images and the grayscale image of the sample image. Then, the principal component analysis method is adopted to compress and reduce the dimensionality of the spliced data. Finally, the image data obtained through principal component analysis is input to SVM, and the image is classified by SVM.

After training the SVM by using a large amount of image data, a trained SVM classifier can be obtained. The test dataset is processed in the same way as the training dataset, and then input to the trained SVM classifier, for observing and analyzing the classification results, which results in the classification strategy.

### 3.1. Feature Extraction

A single feature can only describe a part of the attributes of pictures in a one-sided way. Without the description of distinguishing features, the image classification will not achieve good results. Thus, a single feature can only describe a part of the attributes of pictures in a one-sided way. Without the description of distinguishing features, the image classification will not achieve good results.

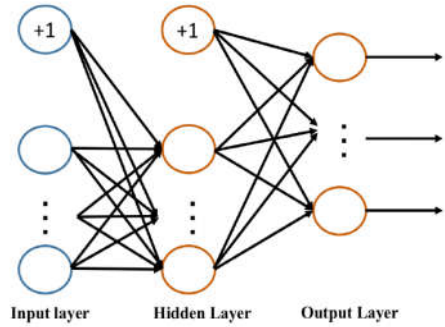

Fig. 2: An Manual Nerve Net Model

Each RGB color image is superimposed by three images corresponding to three views. Our method extract depth characters from each image separately by using a five-layer convolution neural network.

Input Layer:

$$a^{(1)} = x \quad (1)$$

$$(adda\_0^{(1)}) \quad (2)$$

Hidden Layer:

$$z^{(2)} = \theta^{(1)} a^{(1)}; \quad (3)$$

$$a^{(2)} = g(z^{(2)}); \quad (4)$$

$$(adda\_0^{(2)}) \quad (5)$$

Output Layer:

$$z^{(3)} = \theta^{(2)} a^{(2)}; a^{(3)} = g(z^{(3)}) = h_\theta(x); \quad (6)$$

Each layer of a convolution neural network is made up of the two-dimensional plane which has multiple independent neurons. Therefore, a convolution neural network has the function of multi-layer perceptron. C elements of the network are simple elements which make up a convolution layer. S elements of the network are complex elements which make up the down-sampling layer.

Before using the convolutional neural network to extract the feature of the image, it is necessary to pre-process the sample color image, which is necessary to convert the sample color images into R, G and B views of the three components belonging to the map, and then the convolutional neural network is employed to extract R, G and B component image features. The specific algorithm steps are shown below:

**Pre-operation:** separate the sample RGB color image into three images of R, G and B views respectively;

The depth values of R, G and B images are R value, G value and B value, and the depth and the height are the width and the height of the picture respectively.

**Feature Extraction:** using CNN to extract the features of R, G and B images.

**The Convolutional Calculation Method：**

$$conv = \delta\ (imgNet \circ W + b) \qquad (7)$$

**Parameter:**

$\delta$:Activation function   $imgNet$:Image matrix

$\circ$:Convolution operation, W:Convolution kernel, b:Offset value.

**Activation function:**

Sigmoid  $\quad \delta(x) = 1/(1 + e^{-x}) \qquad (8)$

As was said before, the sigmoid function enters a number of real values and then compresses them into a range of 0-1. In particular, large negative numbers are mapped to 0, while large positive numbers are mapped to 1. The sigmoid function has been popular for some time in history because it can well express the meaning of "activation", when 0 is for inactivity and 1 is for full saturation.

**Image matrix**: The picture is input to the computer, through the computer image operation to get the image matrix description. The R component graph can be transformed into a three-dimensional matrix. The three dimensions are the length, width and R value of the image. The G component graph can be converted into a three-dimensional matrix, when the three dimensions are the length, width and G value of the image. The B component graph can be converted into a three-dimensional matrix, when the three dimensions are the length of the image, width and B value.

**Convolution kernel**: Convolution is a commonly used method of image operation. Given an input image, each pixel in the output image is a weighted average of the pixels in a small area of the input image, where weights are defined by a function called a convolution kernel.

**Algorithm steps：**

(1) First convolving the image with a 3 * 3 convolution kernel in the above formula;

(2) Adding b (offset value) to each element of the result (a matrix) obtained in the step (1), and generating N feature maps in the C layer (the value of N can be manually set)

(3) Entering each element in the result (matrix) from step (2) into the activation function, and then getting the S-layer feature map.

(4) According to the number of artificial C and S layers, the above work is carried out in the cycle. In the end, the bottom sampling and the output layer are fully connected, and the final output is obtained.

**Extracting Image Features from Depth Views:**

Here we extracted the depth view of image using the algorithm [18], where a fully convolutional architecture is designed for depth prediction, when endowed with novel up-sampling blocks for dense output maps of the higher resolution. Furthermore, a more efficient scheme is introduced for upconvolutions when combining with the concept of residual learning for creating up-projection blocks and the effective upsampling of feature maps. Finally, the network by optimizing a loss based on the reverse Huber function is adopted for the final depth image prediction.

### 3.2. PCA

Principal component analysis (PCA) is a multivariate statistical analysis method that selects a few important variables by linear transformation of several variables.

We can get the spliced data matrix

$$X = \begin{vmatrix} x_{11} & \cdots & \cdots & x_{1n} \\ x_{21} & \cdots & \cdots & x_{2n} \\ \cdots & \cdots & \cdots & \cdots \\ x_{n1} & \cdots & \cdots & x_{nn} \end{vmatrix} \quad (9)$$

The stitched characterization data is normalized. We get the matrix:

$$A = \begin{vmatrix} a_{11} & \cdots & \cdots & a_{1n} \\ a_{21} & \cdots & \cdots & a_{2n} \\ \cdots & \cdots & \cdots & \cdots \\ a_{n1} & \cdots & \cdots & a_{nn} \end{vmatrix} \quad (10)$$

where

$$a_{ij} = \frac{x_{ij} - \bar{x}_j}{\sqrt{\frac{(x_{ij} - \bar{x}_j)^2}{m}}} \quad i=1,2,3......m; j==1,2,3......n; \quad (11)$$

$$\bar{x}_j = \frac{1}{m} \sum_{i=1}^{m} x_{ij}$$

The correlation matrix is obtained as follows,

$$R = \begin{vmatrix} r_{11} & r_{12} & \cdots & r_{1n} \\ \cdots & \cdots & \cdots & \cdots \\ \cdots & \cdots & \cdots & \cdots \\ r_{n1} & r_{n2} & \cdots & r_{nn} \end{vmatrix} \quad (12)$$

where

$$r_{ij} = \frac{\sum_{i=1}^{m}(a_{ij} - \bar{a}_j)(a_{ik} - \bar{a}_k)}{\sqrt{\sum_{i=1}^{m}(a_{ij} - \bar{a}_j)^2 \sum_{i=1}^{m}(a_{ik} - \bar{a}_k)^2}} \quad (13)$$

i is the specimen number; j,k=1,2,3……n;

$$\overline{a_j} = \frac{1}{n}\sum_{i=1}^{m} a_{ij} \quad (14)$$

This correlation matrix is a symmetric matrix, and thus we take the upper triangle array in the following calculation.

$$R_s = \begin{vmatrix} r_{11} & r_{12} & \cdots & r_{1n} \\ & r_{22} & \cdots & r_{2n} \\ & & \cdots & \cdots \\ 0 & & & r_{nn} \end{vmatrix} \quad (15)$$

Calculate the eigenvalue $\lambda$ and eigenvector $\vec{u}$

Get the component: The obtained eigenvalues are arranged in order of the magnitude $(\lambda_1 > \lambda_2 > ... > \lambda_n)$. Then, w is determined according to the principle ($\frac{\sum_{i=1}^{w} \lambda_i}{\sum_{i=1}^{n} \lambda_i} \geq 85\%$), and the essential presentation which the key principal components are computed are indicated.

### 3.3. SVM

The basic model of the Support Vector Machine (SVM) is to find the optimal separation hyperplane in the feature space so that the positive and negative sample intervals on the training set are maximum. The ω we find is the coefficient of the hyperplane we need. In the field of machine learning, it is a supervised learning model, usually used for pattern recognition, classification and regression analysis. It is based on structural risk minimization theory to construct the optimal hyper-plane segmentation in the feature space, make learning editor to get the global optimization.

Problem description: Assume the training data,

$$(x_1, y_1), ..., \quad (x_1, y_1), x \in R^n, \quad y \in \{+1, -1\}$$

This could be projectioned into a hyper-plane:

$$(\omega \cdot x) + b = 0, \omega \in R^n, b \in R$$

For the normalization:

$$y_i((\omega \cdot x_i) + b) \geq 1, i = 1, ..., l$$

The classification of the interval is equal to: $\frac{2}{\|\omega\|}$, when the maximum interval is equal to the minimum $\|\omega\|^2$.

## 4. Experiments

In order to prove that the proposed method is optimal, we compared the performance of several existing methods in the image classification task.

*4.1. Database*

The database has images for 15 scene categories (such as bedrooms, forests, and office areas), 200 to 400 images per category, and 4,485 images in total.

Caltech256 dataset [17]: this dataset contains 29,780 images of 256 categories with high intra-class and inter-class variabilities. There are at least 80 images for each class of the Caltech-256 dataset.

MIT Indoor dataset [18]: this dataset has 15,620 images of 67 indoor scenes, when it is more difficult to classify than the Scene15 dataset not only because it has more classes but also because the intra-class variation of the MIT Indoor dataset is relatively larger.

We set the same experimental conditions: extract SIFT descriptors on overlapping pixels with an overlap of 6 pixels. The local area is reshaped to 16*16 pixels. Sparse coding with locality constraint [19] is employed to encode local features along with max pooling for image feature learning. The spatial pyramid structure with three different scales is used to fuse the spatial relations of local features. We set the codebook size of 1024 for these three data sets. As with other methods, we randomly select training images to ensure reliable results. We use SVM classifiers with different loss functions for semantic spatial structure and image classification prediction. Performance evaluation is based on the average classification rate of every class.

Table 1 The Results of Different Approaches on the Scene15 database

| Method | Accuracy (%) |
| --- | --- |
| K-SPM [16] | 81.40±0.50 |
| KC-SPM [20] | 76.70±0.40 |
| K-SPM[21] | 76.73±0.65 |
| Sparsecoding-SPM[21] | 80.28±0.93 |
| ObjectBank[22] | 80.9 |
| Low-dimensional Semantic Spaces [23] | 72.20±0.20 |
| Ours without depth view | 81.06±0.24 |
| Ours | 85.33±0.47 |

*4.2. Results on the Scene15 Database*

The performance comparison of between our method and several existing methods [16, 20, 21 ,22, 23] is based on the Scene15 dataset [16] shown in table 1. The upshot is that: (1) Our method performs much better than sparse coding [21] by about 5.05%, which justified that using high-level image features can help image classification better than using visual features directly. (2) Our method, compared with other semantic-based methods, can obtain more descriptive information for classifier.

### 4.3. Results on the Caltech256 Database

The performance comparison of between our method and several existing methods [6, 20, 21, 22, 23, 24] is based on the Scene15 dataset [17] shown in table 2. The similar conclusions can be fined on the Scene15 dataset. The proposed method outperforms that of local features with sparse coding [16, 20, 21] or its variants [24]. Our method also has the better performance than other semantic-based methods [22, 23] by extracting more efficient high-level features. Experimental comparisons with other methods show the efficiency of the proposed model.

Table 2 The Results of several approach about Caltech 256 Dataset

| Algorithm | 15Training | 30Traning |
| --- | --- | --- |
| K-SPM[16] | 28.37±0.53 | 31.24±0.58 |
| KC-SPM[20] | 21.63±0.47 | 27.27±0.46 |
| K-SPM[21] | 27.72±0.51 | 29.51±0.52 |
| Sparsecoding-SPM [21] | 27.73±0.51 | 34.02±0.35 |
| ObjectBank [22] | 32.13 | 39.00 |
| Low-dimensional Semantic Spaces [23] | 30.14±0.34 | 37.20±0.23 |
| Classemes[24] | 30.67 | 36.00 |
| Ours without depth view | 28.32±0.23 | 30.67±0.19 |
| Ours | 39.56±0.58 | 45.01±0.51 |

### 4.4. Comparisons on the MIT Indoor Dataset

Table 3 shows the performance comparison of the pro- posed method on the MIT Indoor dataset [18] with the popular image classification approaches [21, 22].

Table 3 The Comparison of Different Methods on the MIT Indoor Dataset

| Algorithm | Accuracy (%) |
| --- | --- |
| Sparsecoding-SPM [21] | 24.37±035 |
| ObjectBank [22] | 37.6 |
| Classemes [24] | 26.02 |
| Ours without depth view | 32.45±0.16 |

The proposed method in this paper achieves a classification performance of 42.18% on the MIT Indoor dataset which outperforms the other semantic-based methods, such as OB (37.6%) [22] and Classemes (26.02%) [24]. Since the MIT Indoor dataset has more scene classes than the Scene15 dataset [16], it is more difficult to be classified. The intra-class variations are relatively larger.

## 5. Conclusion

In this paper, an image classification technology is proposed by making principal component analysis on multi-view deep features. In detail, firstly, the depth image of the original image is estimated; secondly, deep features are extracted from both the RGB views and the depth view Separately; thirdly the reducing dimension operation is implemented with the PCA; Eventually, the SVM is adopted for image classification.